\documentclass[conference]{IEEEtran}

\IEEEoverridecommandlockouts
\usepackage{cite}
\usepackage{amsmath,amssymb,amsfonts}
\usepackage{algorithmic}
\usepackage{setspace}
\usepackage{graphicx}
\usepackage{textcomp}
\usepackage{xcolor}
\usepackage{float}
\usepackage{url}
\usepackage{multirow}

\usepackage{xcolor,colortbl}

\definecolor{aliceblue}{rgb}{0.94, 0.97, 1.0}

\definecolor{LightCyan}{rgb}{246,246,174}

\usepackage{commath}
\usepackage{ulem}

\usepackage{hyperref}
\hypersetup{
    colorlinks=true,
    linkcolor=blue,
    filecolor=magenta,      
    urlcolor=cyan,
}

\graphicspath{ {./} }

\def\BibTeX{{\rm B\kern-.05em{\sc i\kern-.025em b}\kern-.08em
    T\kern-.1667em\lower.7ex\hbox{E}\kern-.125emX}}
\begin{document}

\title{ 2D versus 3D Convolutional Spiking Neural Networks Trained with Unsupervised STDP\\for Human Action Recognition }
\author{
\IEEEauthorblockN{
Mireille El-Assal\IEEEauthorrefmark{1},
Pierre Tirilly\IEEEauthorrefmark{1},
and Ioan Marius Bilasco\IEEEauthorrefmark{1}
}
\IEEEauthorblockA{\IEEEauthorrefmark{1}
\textit{Univ. Lille, CNRS, Centrale Lille,}
\textit{UMR 9189 -- CRIStAL -- Centre de Recherche en Informatique, Signal et Automatique de Lille}\\
F-59000, Lille, France\\
}
\IEEEauthorblockA{
Email: mireille.elassal2@univ-lille.fr, pierre.tirilly@univ-lille.fr, marius.bilasco@univ-lille.fr
}
}
\maketitle  

\begin{abstract}
Current advances in technology have highlighted the importance of video analysis in the domain of computer vision. However, video analysis has considerably high computational costs with traditional artificial neural networks (ANNs). Spiking neural networks (SNNs) are third generation biologically plausible models that process the information in the form of spikes. Unsupervised learning with SNNs using the spike timing dependent plasticity (STDP) rule has the potential to overcome some bottlenecks of regular artificial neural networks, but STDP-based SNNs are still immature and their performance is far behind that of ANNs. In this work, we study the performance of SNNs when challenged with the task of human action recognition, because this task has many real-time applications in computer vision, such as video surveillance. In this paper we introduce a multi-layered 3D convolutional SNN model trained with unsupervised STDP. We compare the performance of this model to those of a 2D STDP-based SNN when challenged with the KTH and Weizmann datasets. We also compare single-layer and multi-layer versions of these models in order to get an accurate assessment of their performance. We show that STDP-based convolutional SNNs can learn motion patterns using 3D kernels, thus enabling motion-based recognition from videos. Finally, we give evidence that 3D convolution is superior to 2D convolution with STDP-based SNNs, especially when dealing with long video sequences.
\end{abstract}

\begin{IEEEkeywords}
spiking neural networks, convolution, spatio-temporal, action classification, STDP, unsupervised learning.
\end{IEEEkeywords}

\section{Introduction}
Computer vision is a continuously growing field in machine learning, and video analysis is one of the major challenges in this field. However, deep neural networks, that are currently the state-of-the-art method in video analysis, have limitations in terms of computational costs and need for labeled data. On the other hand, models like spiking neural networks (SNNs) trained with unsupervised spike timing-dependent plasticity (STDP) can reduce the need for labeled data by learning visual features in an unsupervised fashion \cite{ImprSNNTrain}. SNNs are models that process and transmit the information in the form of low-energy spike trains, similarly to the way information is processed in the human brain. These spikes are sparse in time, which means that they can potentially hold a large amount of information \cite{b8}. Furthermore, SNNs perform local computations, which promotes easier implementation with ultra-low-power neuromorphic hardware, in addition to parallelizing the training process, hence speeding it up. However, despite all of these advantages, these models have limitations of their own, such as frequency loss \cite{MasteringOutputFrequency}, and the subsequent difficulty of training deep SNN architectures without supervision \cite{multLyrSNNWithTargetTmStampTrshAdpt}. Video analysis with SNNs is still a relatively new topic. One major challenge is to find a cost-friendly method that needs little to no pre-processing and is able to extract relevant spatio-temporal features from videos. In this work, we tackle the task of human action recognition using unsupervised STDP-based 3D convolutional SNNs. This type of learning, with 3D convolutional SNNs that learn spatio-temporal information found in videos by sliding their convolutional kernels in the temporal dimension, is still not explored to the best of our knowledge. However, this is in theory a processing-cost-friendly method that could be used in real-world applications. Therefore, it is interesting, yet challenging, to implement 3D convolutions in the spiking domain with unsupervised learning.

This paper introduces unsupervised spatio-temporal feature learning with 3D convolution using STDP-based SNNs. This 3D convolutional SNN is used to extract spatio-temporal features from human action recognition videos. We compare this model to its 2D equivalent in order to reach an accurate assessment of the performance of these STDP-based SNNs by listing the benefits and drawbacks of each method. Experiments are performed on the \href{https://www.csc.kth.se/cvap/actions/}{KTH} \cite{b39} and \href{http://www.wisdom.weizmann.ac.il/~vision/SpaceTimeActions.html}{Weizmann} \cite{b37} datasets, which are simple, early datasets that could be compared to MNIST, but for action recognition. Although very high recognition rates have already been achieved on these datasets using traditional computer vision approaches \cite{KTHGoodPErcent}, their simplicity makes them good basic benchmarks to study the performance of new models like the ones targeted in this paper. Feature extraction with 3D convolution on STDP-based SNNs is a first step towards bridging the performance gap between SNNs and other deep learning solutions in complex vision tasks. This work can serve as an interesting building block in developing 3D spiking models that can learn spatio-temporal characteristics and perform their training locally. The main contributions of this paper are summarized as follows: 
\begin{itemize} 
  \item we present a spiking model for 3D convolution that allows learning spatio-temporal patterns with STDP in an unsupervised manner;
  \item we include this 3D convolution model into a state-of-the art spiking architecture for unsupervised feature learning;
  \item we evaluate and compare the performance of 2D and 3D SNNs on both of the KTH and Weizmann action recognition datasets as raw videos and as pure motion information;
  \item we give an analysis of the effects of the main hyper-parameters on the performance of a 3D convolutional SNN.
\end{itemize}

\section{Related Work}
\noindent \textbf{Convolutional neural network architectures.} There are many popular convolutional neural networks in the literature, such as AlexNet \cite{AlexNet}, VGGNet \cite{VggNet}, GoogleNet \cite{GoogleNet}, and ResNet \cite{ResNet}. These models are 2D architectures used mostly for image classification, and there is some work that presents convolutional SNNs inspired by these models \cite{SNNVggandResNet}. Convolutional neural networks are often used with both image and video data. However, with video analysis, 2D CNNs process video frames one at a time, and usually need an extra processing step to make sense of the motion information between the frames. Meanwhile, 3D CNNs can naturally extract spatio-temporal features from videos by leveraging its temporal dimension. Therefore, 3D CNNs are good candidates when it comes to video analysis, either in classification~\cite{kth3d9frames} or in regression tasks~\cite{videobasedfacealRomain}. In \cite{LearningSPTFeatures}, the authors handle the task of spatio-temporal feature learning using deep 3D CNNs and compare them to 2D CNNs. They conclude that 3D models are better suited for video analysis tasks than 2D models. In \cite{kth3d11frames}, the authors use a 3D motion cuboid for action detection and recognition; they use 11 frames as inputs, and they reach classification rates of 94.9\% with the KTH dataset, and 97.2\% with the Weizmann dataset. In \cite{kth3d9frames}, the authors present another 3D CNN model for action recognition. This model captures the motion information encoded in multiple adjacent frames (9-frame input) using 3D convolutions. This model generates multiple channels of information from the input frames, and the final feature representation is obtained by combining information from all channels. They achieve a classification rate of 90.2\% with the KTH dataset.

However, all of the 3D CNNs mentioned in this section use traditional analog values, and not spikes. In our work, we transpose 3D convolution to be used with SNNs in an unsupervised manner, in order to decrease the computational and labeling costs.

\noindent \textbf{Convolutional spiking neural network architectures for spatio-temporal information learning.}
SNNs are used for video analysis tasks, especially with the emergence of dynamic vision sensors \cite{EventBasedVision}. In \cite{DSNNMotionSeg}, the authors create a deep encoder-decoder SNN architecture for motion segmentation, where they take the input from a DVS camera, and they use back-propagation with their own spatio-temporal loss function. ANN-to-SNN transformation is applied in \cite{EFFProcessSTDataStream}, where the authors train a regular ANN for a given sequence of input frames. They use streaming rollouts to allow temporal integration over multiple frames. However, they do not address the ANN bottlenecks that SNNs can avoid, because they use a regular ANN to conduct the training with supervised learning. A supervised approach is also suggested in \cite{STBPHighPerformenceSNN}, where the authors use a supervised spatio-temporal back-propagation (STBP) algorithm for training SNNs and a loss function that includes the mean square error for all samples under a given time window. Another SNN learning method is the BCM (Bienenstock, Cooper, and Munro) learning rule. In \cite{HARSNNGRN}, the authors proposed a BCM-based SNN model that classifies human action recognition videos. Another learning rule is STDP, which is a biologically plausible unsupervised learning rule \cite{ImprSNNTrain}. In \cite{b24}, the authors use Reward-modulated Spike Timing-Dependent Plasticity (R-STDP) and reinforcement learning to train their network to perform action classification. In \cite{elassal:hal-03263914}, the authors use a 2D convolutional STDP-based SNN to process videos. However, they use costly pre-processing to conserve the temporal information between the video frames, and motion is only processed through these non-neuromorphic preprocessing steps. In \cite{oudjail:hal-02529895}, the authors present an interesting study of the impact of varying the meta-parameters of an STDP-based SNN on the task of motion recognition, but they only use a small range of spiking pixels in their experiments, and thus they do not test their network in real-world scenarios. There is very little work in the literature that concerns convolutional SNNs with convolutional kernels that slide in the temporal domain. In \cite{zhang2021tuning}, the authors train a multi-layer SNN using their own reward propagation algorithm. They use 1D and 2D convolutions for the temporal and spatial domains respectively. In \cite{spyketorch}, the authors consider an extra dimension in their tensor to represent time, but they separate input samples into bins of time. 

In our work, the convolutional kernels slide in the spatial and temporal dimensions simultaneously, and not separately. Moreover, our input samples are taken as tensors that have a time dimension, but without further pre-processing. 

\noindent \textbf{3D convolutional spiking neural network architectures.}
There is little work in the literature that addresses the issue of using 3D convolution for video analysis with SNNs. In~\cite{newSCRNN}, the authors combine 3D convolutional SNNs with recurrent neural networks (RNNs), and they use a supervised Spike Layer Error Reassignment (SLAYER) training mechanism to train their network. In \cite{doborjehSNNcube3D}, the authors use an SNN cube as a 3D brain-like structure with recurrent connections. However, the type of data they are interested in analysing is Spatio-Temporal Brain Data (STBD), that is used in the medical domain, such as EEG and fMRI, not video data. 

In this work, we propose 3D convolution for training an SNN to learn spatio-temporal features used for action classification. We use the biological STDP learning rule \cite{ImprSNNTrain} in an unsupervised manner to train our convolutional SNN to extract spatio-temporal features. This unsupervised learning has the advantage of not needing labeled data. We explore the implementation compatibility of mechanisms used to train 2D SNNs with a multi-layer 3D SNN architecture.

\section{Spatio-temporal feature learning with STDP}
\subsection{Baseline Architecture}
Implementing a multi-layer SNN is a very challenging task \cite{multLyrSNNWithTargetTmStampTrshAdpt}, in part as a consequence of the frequency loss problem. This problem results from the fact that multiple input spikes are needed to cross the membrane potential of a spiking neuron, prompting it to fire a single output spike. To avoid this problem, we rely on the general, state-of-the-art, convolutional spiking neural network architecture from \cite{ImprSNNTrain}. It enables multi-layer training thanks to its choice of neural coding, homeostasis rule, and training protocol.

This model consists of feed-forward layers that contain integrate-and-fire (IF) neurons \cite{IFneuron}. The IF neuron model is characterised by having a certain membrane potential $v(t)$ and a threshold potential $v_{\text{th}}(t)$. Input spikes are integrated into the neuron membrane potential, until it reaches the threshold potential, therefore triggering the neuron to fire an output spike. Then, the membrane potential is reset to its resting potential $v_{r}$, which is $0$ volts in this work. The behaviour of this neuron is characterised by the following equations \cite{ImprSNNTrain}:
\begin{equation}
\begin{array}{c}
\displaystyle C_{m} \frac{\partial v(t)}{\partial t} = \sum_{i \in \mathcal{E}} v_{i} f_{s}(t-t_{i})\\ v(t) \leftarrow v_{r} \mbox{ when }  v(t) \geq v_{\text{th}}(t) 
\end{array}
\end{equation}
\begin{equation}
f_{s}(x) = 
\begin{cases}
    1, & \mbox{if } x \geq 0 \\ 
    0, & \mbox{otherwise}
\end{cases}
\end{equation}
where $C_{m}$ is the membrane capacitance, $\mathcal{E}$ is the set of incoming spikes, $v_{i}$ is the spike voltage of the \(i\)-th spike, $t_{i}$ is the timestamp of the  \(i\)-th spike, and $f_{s}$ is the kernel of spikes. The training is unsupervised using the biological STDP learning rule, which provides better performance than additive STDP and multiplicative STDP, by adding non-linearity, as explained in \cite{ImprSNNTrain}. It allows the learning of more complex features. This learning rule is characterised by the following equation \cite{ImprSNNTrain}:

\begin{equation}
\Delta_{w} = 
\begin{cases}
    \eta_{w} e^{-\frac{t_{ \text{pre}} - t_{\text{post}}}{\tau_{\text{STDP}}}}, & \mbox{if } t_{\text{pre}} \leq t_{\text{post}} \\ 
    -\eta_{w} e^{-\frac{t_{\text{pre}} - t_{\text{post}}}{\tau_{\text{STDP}}}}, &  \text{otherwise}
\end{cases}
\end{equation}

where $\eta_{w}$ is the learning rate and $\tau_{\text{STDP}}$ is the time constant that controls the STDP learning window, and $t_{\text{pre}}$ and $t_{\text{post}}$ are the firing timestamps for input and output neurons respectively.

The network uses winner-takes-all (WTA) inhibition to prevent several neurons from learning the same pattern. With WTA, some neurons can overpower other neurons, i.e. they have a tendency to fire more spikes than others. This leads to the network becoming stuck in a state where a few active neurons fire all the time, while the other neurons are quiet. In order to ensure the stability of the network, a homeostasis mechanism is needed. We use the threshold adaptation method introduced in \cite{multLyrSNNWithTargetTmStampTrshAdpt}. This method trains the neurons to fire at a given objective time $t_{\text{obj}}$ to maintain the homeostasis of the network. Within this method, each time a neuron fires or receives an inhibitory spike, the thresholds of all neurons (winners and losers) are adapted so that their firing time converges towards $t_{\text{obj}}$. These thresholds are updated according to equations (4), (5), and (6).
\begin{equation} 
\Delta_{\text{th}}^1 = -\eta_{\text{th}}(t - t_{\text{obj}})
\end{equation}

\begin{equation}
\Delta_{\text{th}}^2 = 
\begin{cases}
    \eta_{\text{th}}, & \mbox{if } \mbox{$t_{i} = \mathrm{min}(t_{0},...,t_{N})$} \\ 
    -\frac{\eta_{\text{th}}}{l_{d}(n)}, & \mbox{otherwise}
\end{cases}
\end{equation}

\begin{equation} 
v_{\text{th}}(t) = \mathrm{max}(\text{th}_{\mathrm{min}}, v_{\text{th}}(t-1)+ \Delta_{\text{th}}^1 + \Delta_{\text{th}}^2) 
\end{equation}

where $t$ is the timestamp at which the neuron fires, $\eta_{\text{th}}$ is the threshold learning rate, $l_{d}$ is the number of neurons in competition in the layer, $t_{i}$ is the firing timestamp of neuron $i$, and $\text{th}_{\mathrm{min}}$ is the minimum possible threshold value \cite{ImprSNNTrain}.

\begin{figure}
\centerline{\includegraphics[scale=0.2]{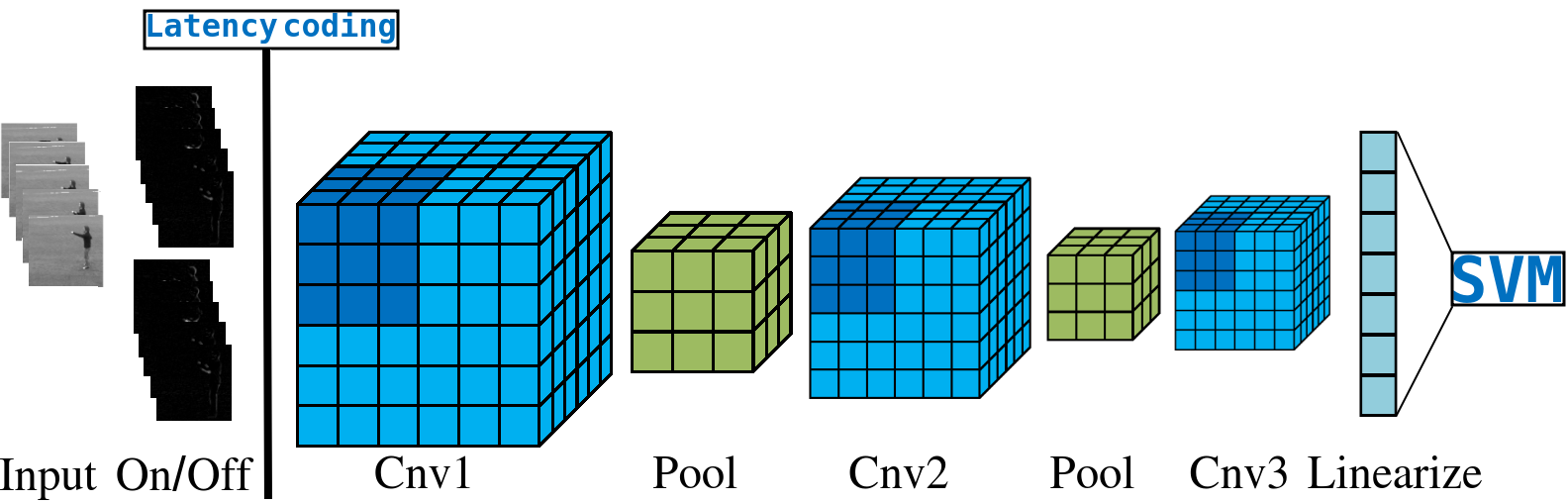}}
\caption{Network topology.}
\label{fig1}
\end{figure}

The multi-layer SNNs we use are made up of convolution and max-pooling layers, as shown in Figure \ref{fig1}. The output of this network is then linearized using sum pooling. This output is introduced into a support vector machine (SVM) with a linear kernel, which performs the action classification. An SVM is used to classify the samples because we focus on the unsupervised learning of features. Any other supervised method can be used for the final classification; we chose an SVM because it is standard and effective with default hyper-parameters.

\subsection{Video Representation}
A video is a sequence of image frames, where the action is perceived by the displacement of pixels that represent the objects in motion between each two consecutive frames. An image is a 3D tensor of size $w \times h \times c$ where $w$ is its width, $h$ its height, and $c$ its number of channels. Therefore, a video can be represented as a 4D tensor of size $w \times h \times c \times td$ where $td$ is the temporal depth, which is equivalent to the number of video frames. These tensors contain continuous values that represent the pixels, and they are transformed using on-center/off-center coding and temporal coding (with one spike per pixel) into tensors of timestamps that have coordinates and represent spikes. In the case of image data, these spikes have coordinates $(x, y, k)$, where $x$ and $y$ are the spatial coordinates, and $k$ is the channel coordinate. In the case of video data, these coordinates are $(x, y, z, k)$, where $z$ is the temporal coordinate.

\subsection{2D convolution}
\begin{figure}
\centerline{\includegraphics[scale=0.2]{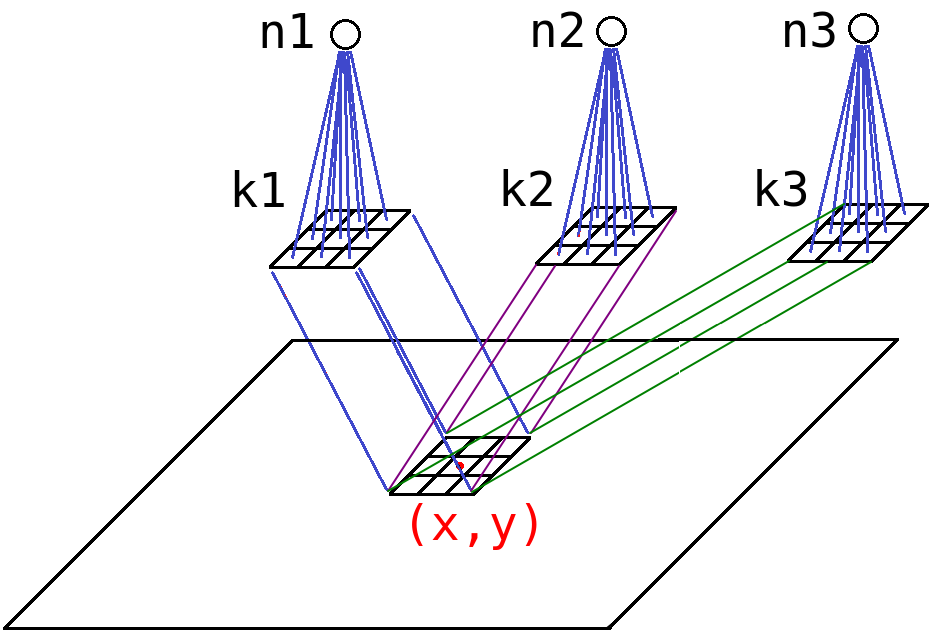}}
\caption{The neighborhood of an output neuron at \((x, y)\) is a plane, and n1, n2, and n3 are competing neurons at the same location. The channel dimension is not drawn in this figure in order not to confuse it with the temporal dimension in Figure \ref{fig:3DRS}.}
\label{fig:2DRS}
\end{figure}

Convolution layers preserve the shape and spatial structure of the input, and they allow the reduction of the number of trainable parameters (i.e. synaptic weights) when using shared weights. In opposition to dense layers, neurons in convolution layers are connected only to a subset of the neurons of the previous layer. A 2D convolutional layer is naturally used for spatial feature extraction. It has a set of $k$ trainable filters of size $f_{w} \times f_{h}$. Each neuron is connected to $f_{w} \times f_{h}$ neurons of the previous layer. These convolutional filters slide across the spatial dimensions of an input of size $l_{w} \times l_{h}$ with a given stride (step between two consecutive locations). Figure~\ref{fig:2DRS} illustrates this process. Formally, a spiking 2D convolution operator can be described as follows:
\begin{equation}
v_{x,y,k}(t) = \sum_{n \in \mathcal{N}} W_{i(x_{n}), j(y_{n}), k_{n}, k} \times f_{s}(t-t_{n})
\end{equation}
where $v(t)$ is the neuron membrane potential at time $(t)$, $x, y$ and $k$ are the coordinates of the neuron in the width, height, and channel dimensions, respectively, $W$ is the trainable synaptic weight matrix, $i()$ and $j()$ are functions that map the location of the input neuron to the corresponding location in the weight matrix, and $k$ is the index of the trainable filter. When $v_{x,y,k}(t)$ crosses the $v_{\text{th}}(t)$ of the IF neuron, this neuron fires an output spike. During the training phase, it triggers the STDP learning rule. In this case, all the weights and thresholds of the layer are updated according to Equations (3) and (6) in Section III-A. In practice, for each input sample, we only select a certain number $n_{\text{sampling}}$ of spatial locations at which neurons processed. It prevents the same filter from being updated at many locations in parallel, improving the convergence of the network. We calculate $n_{\text{sampling}}$ based on the sizes of the input sample and of the convolutional filter, as shown in Equation (8):
\begin{equation}
n_{\text{sampling}} = \frac{2 \times l_{w} \times l_{h}}{f_{w} \times f_{h}}
\end{equation}
The network is trained layer-wise, i.e. each layer is trained independently, starting from the first one, then its weights are freezed during the training of the subsequent layers. Neuron sampling occurs only in the layer being currently trained. During the testing phase, all neurons are active and layers are processed sequentially, i.e. all the input spikes of one layer are processed before processing the next layer.

\begin{figure}
\centerline{\includegraphics[scale=0.35]{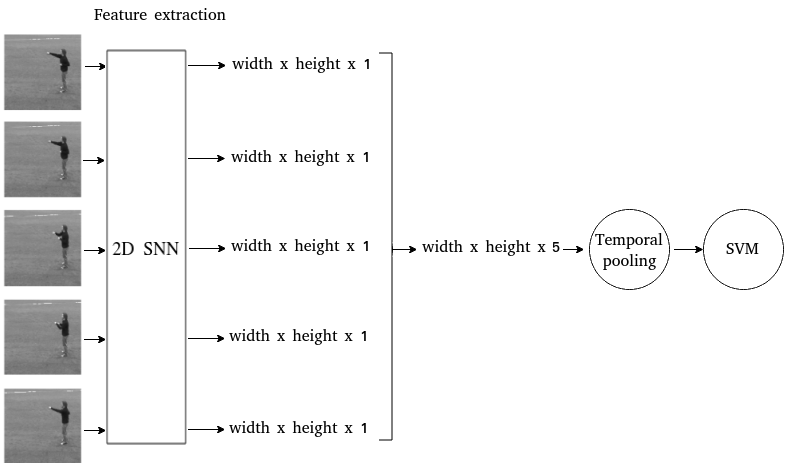}}
\caption{Spatio-temporal feature extraction with a 2D architecture.}
\label{fig:2DArch}
\end{figure}

Since 2D convolutions only process the spatial dimensions of the input, they ignore any temporal information. Therefore, spatio-temporal feature extraction with such architectures must be achieved by implementing certain methods that conserve the temporal component of the input information, like the several methods mentioned in Section II. In this work, we conserve the temporal component of the input videos by processing each frame separately and then using temporal sum pooling to join the extracted features before using the SVM for sample classification, as shown in Figure \ref{fig:2DArch}. It allows us to focus on the features learned by 2D convolutions only and compare them with the ones learned by 3D convolutions.

\subsection{3D convolution}
\begin{figure}
\centerline{\includegraphics[scale=0.15]{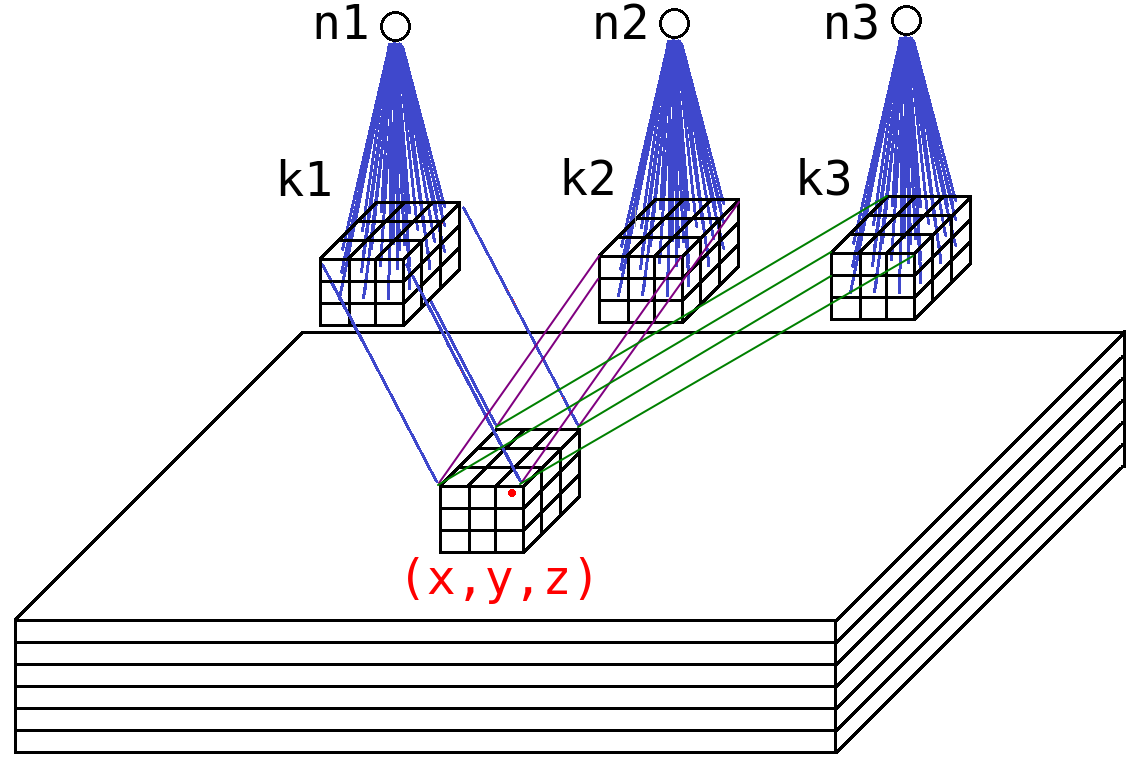}}
\caption{The neighbourhood of a neuron at \((x, y, z)\) is a volume, and n1, n2, and n3 are competing neurons at the same location.}
\label{fig:3DRS}
\end{figure}

The filters of a 3D convolutional SNN can slide along the temporal dimension of the input tensor in addition to the spatial ones, and extract spatio-temporal features that correspond to the movement occurring in the input video. The virtual movement of the 3D convolutional filters can be visualised as going through the three dimensions, width, height, and temporal depth, in steps determined by the stride in each dimension. Each neuron in the convolution output processes a part of the data sample in both space and time. A 3D convolution layer is therefore defined by a set of $k$ trainable filters, with sizes $f_{w} \times f_{h} \times f_{td}$, where $td$ stands for temporal dimension. As in 2D, but with the extra dimension, each neuron of a layer is connected to $f_{w} \times f_{h} \times f_{td}$ neurons of the previous layer. The coordinates of a neuron or a spike in this 3D model are now $x, y, z$, and $k$. Figure~\ref{fig:3DRS} illustrates this process. 3D spiking convolution can be formalized as: 
\begin{equation}
v_{x,y,z,k}(t) =\sum_{n \in \mathcal{N}} W_{i(x_{n}), j(y_{n}), m(z_{n}), k_{n}, k} \times f_{s}(t-t_{n})
\end{equation}
where the matrix of trainable synaptic weights $W$ has an additional temporal dimension, so $m()$ is a function that maps the temporal location of the input neuron to the corresponding temporal location in the weight matrix, and $z$ is the temporal coordinate of the selected neuron. Similarly to Section III-C, when the membrane potential $v_{x,y,z,k}(t)$ crosses the threshold potential $v_{\text{th}}(t)$, Equations (3) and (6) are applied to update the synaptic weights and thresholds of the network. It is important to note that the threshold adaptation rule and the biological STDP rule are the same in both 2D and 3D architectures as they are independent of the input and filter dimensions. The training and testing processes are then the same as in the case of 2D convolution. We only need to update Equation (8) to account for the temporal dimension:
\begin{equation}
n_{\text{sampling}} = \frac{3 \times l_{w} \times l_{h} \times l_{td}}{f_{w} \times f_{h} \times f_{td}}
\end{equation}

\begin{figure}
\centerline{\includegraphics[scale=0.4]{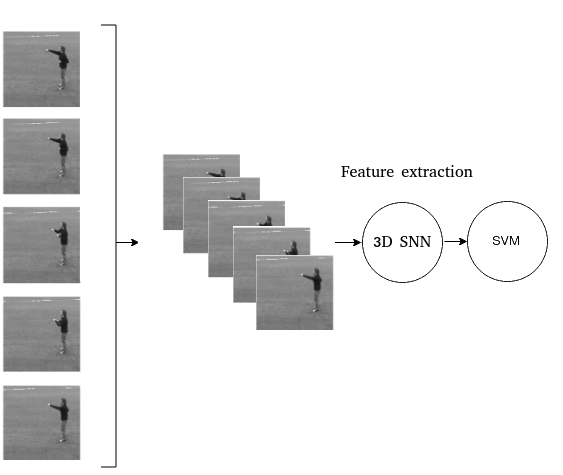}}
\caption{Spatio-temporal feature extraction with a 3D architecture.}
\label{fig:3DArch}
\end{figure}

The 3D convolutional network takes a video sample as input, in the form of a 4D tensor where the temporal depth corresponds to the number of frames per video, as shown in Figure \ref{fig:3DArch}. Therefore, an input stack of frames of size $l_{w} \times l_{h} \times l_{td}$ can be processed naturally by a 3D convolutional SNN. Extra processing steps like optical flow extraction and the fusion of output features are not needed with such models.

\section{Evaluation}
This section contains the details of our experiments. First, we investigate the hyper-parameters of the model that impact the most the features to be learned: \(t_{\mathrm{obj}}\), the kernel size, and the video length. Then, we analyze the properties of the spatial and spatio-temporal features learned with 2D and 3D convolutions, respectively.

\subsection{Datasets and evaluation protocol}
The KTH dataset contains 600 videos made up of 25 subjects performing 6 actions in 4 scenarios. The subjects 11, 12, 13, 14, 15, 16, 17 and 18 are used for training, while 19, 20, 21, 23, 24, 25, 01, 04 are used for validation and 02, 03, 05, 06, 07, 08, 09, 10 and 22 are used for testing, as indicated in the KTH protocol. To shorten the running time of experiments, we take a subset of the KTH video frames (like in \cite{kth3d11frames} and \cite{kth3d9frames}). We used different temporal sizes in our experiments, however we only report the experiments with $8$ and $20$ frames per video. This is because $8$ frames has proven to be the most suitable video length with 2D SNNs. We use a temporal size of $20$ to compare the performance of 2D SNNs versus 3D SNNs with longer sequences. We skip one frame between each two consecutive frames in order to make sure to capture a full cycle of the performed action. The Weizmann dataset contains 90 videos of 9 subjects performing 10 actions. The experiments on this dataset are all done using the leave-one-out strategy. We sample the Weizmann video frames in the same way as those of the KTH dataset. We also scale down the frame sizes of both datasets to half of their original sizes for processing speed reasons. We measure the classification accuracy (in \%) on the test set for all experiments. Each experiment was run three times and we report the average accuracy over the three runs. 

\subsection{Pre-processing}
The videos are presented to the SNNs either as raw frames in some experiments, or pre-processed with background subtraction in other experiments. This method consists of subtracting each two consecutive frames to remove the static spatial information. We use this method in order to see the ability of the SNNs to classify input information that consists purely of motion. In the rest of this section, experiments are performed on raw frames unless otherwise specified.

\subsection{Implementation details}
The meta-parameters used in this work are presented in Table \ref{table:I}. A difference-of-Gaussian (DoG) filter is used to simulate on-center/off-center cells. Experiments with and without this filter were conducted, and experiments without this filter gave inferior results. The DoG filter has a kernel of size \(\mathrm{DoG}_{\mathrm{size}} = 7\), and uses centered Gaussians of variance \(\mathrm{DoG}_{\mathrm{in}} = 1.0\) and \(\mathrm{DoG}_{\mathrm{out}} = 4.0\). No padding is used for the convolutions. The convolutional filters of 2D layers use a stride of $1$ in all dimensions, while those of 3D layers use a stride of $1$ in spatial dimensions, and a stride of $2$ in the temporal dimension. The max pooling layers use a kernel of size $2 \times 2$ and a stride of $2 \times 2$ for the 2D setup, and a kernel of size $2 \times 2 \times 2$ and a stride of $2 \times 2 \times 2$ for the 3D setup.

The convolutional SNNs tested in this work are simulated using the \href{https://gitlab.univ-lille.fr/bioinsp/falez-csnn-simulator/tree/07fd14324afc42d7b3b24a3472271e1c6a90255a}{csnn-simulator} \cite{ImprSNNTrain}, which is open-source and publicly available. The source code for our experiments will be released publicly.

\begin{table}
\begin{center}
 \begin{tabular}{| c |} 
 \hline
   \rowcolor{aliceblue}\textbf{STDP} \\
 \hline
 $\eta_{w}= 0.1$, $\tau_{\mathrm{STDP}}= 0.1$, $W \sim U(0, 1)$ \\
 \hline
 \rowcolor{aliceblue}\textbf{Threshold Adaptation }\\
 \hline
 $t_{\mathrm{obj_{KTH}}}= 0.65, 0.3, 0.1$,  $\mathrm{th}_{\mathrm{min}}= 1.0 $, $\eta_{\text{th}}= 1.0$,\\ $t_{\mathrm{obj_{Weizmann}}}= 0.75, 0.55, 0.15$, $\upsilon_{\text{th}}(0) \sim G(5, 1)$ \\
 \hline
 \rowcolor{aliceblue}\textbf{Difference-of-Gaussian} \\
 \hline
 $\mathrm{DoG_{in}}= 1.0$, $\mathrm{DoG_{out}}= 4.0$, $\mathrm{DoG_{size}}= 7.0$ \\
 \hline
\end{tabular}
\end{center}
\caption{The meta-parameter values used in the experiments.}
\label{table:I}
\end{table}

\subsection{Expected Timestamp}
The threshold adaptation method of Section III-A requires an objective timestamp $t_{\text{obj}}$ towards which the firing time of the neuron must converge. This value can vary with different types of input information, and is selected for a given dataset using trial and error. This hyper-parameter has a strong impact on the nature and quality of the features learned by the network \cite{multLyrSNNWithTargetTmStampTrshAdpt}, so it is important to find the most suitable value. Training 2D and 3D single-layer architectures with raw videos as an input gives the results displayed in Table \ref{table:obbjtime} with filter sizes of $5 \times 5$ and $5 \times 5 \times 2$ for the 2D and 3D SNNs respectively. These results show the effect of spike selectivity on the resulting classification rate, where small and large values of objective times gives similar classification rates with single layer 2D and 3D architectures. Therefore, promoting a high spike selectivity gives very similar results to those obtained when integrating more spikes. 

In the case of both 2D and 3D multi-layer SNNs, high selectivity of spikes in previous layers (by choosing a small value for the objective time) degrades the learning in the next layers, because not enough spikes can be integrated. Therefore, the objective time needs to be large enough to promote activity. However, a high $t_{\text{obj}}$ in all of the layers gives inconsistent performance with the KTH dataset, and decreases classification rates in subsequent layers, as shown in Table \ref{table:obbjtimesimilar}. Increasing the selectivity over layers (by choosing decreasing objective times) can consistently increase the classification rates throughout the layers. Therefore, for the rest of the experimental procedure, the $t_{\text{obj}}$ values for the KTH dataset will be $0.65, 0.3, 0.1$ for the three layers of the 2D and 3D architectures. Similarly, the $t_{\text{obj}}$ values for the Weizmann dataset will be $0.75, 0.55, 0.15$ for the three layers of the 2D and 3D architectures. 

\begin{table}
\begin{center}
\begin{tabular}{|c | c | c | c | c |}  
\hline
\rowcolor{aliceblue}Dataset  & \multicolumn{2}{c|}{KTH} & \multicolumn{2}{c|}{Weizmann} \\
\hline
\rowcolor{aliceblue} Objective time & 0.1 & 0.65 & 0.15 & 0.75 \\ [0.5ex] 
\hline
$2D$ & 56.84 & 56.94 & 47.26 & 48.63  \\ [0.5ex] 
\hline
$3D$ & 55.09 & 57.56  & 51.45 &  49.49 \\ [0.5ex] 
\hline
\end{tabular}
\end{center}
\caption{Classification rates in \% of KTH and Weizmann dataset (8-frame videos) over 3 runs, as a function of different objective time values with 2D and 3D single-layer architectures.}
\label{table:obbjtime}
\end{table}

\begin{table}
\begin{center}
\begin{tabular}{|c | c | c | c | c | c | c |}  
\hline
\rowcolor{aliceblue}   & \multicolumn{3}{c|}{$t_{\text{obj}}$ = 0.65} & \multicolumn{3}{c|}{$t_{\text{obj}}$ = 0.65 0.3 0.1}\\
\hline
\rowcolor{aliceblue}Layers & Cnv1 & Cnv2 & Cnv3 & Cnv1 & Cnv2 & Cnv3 \\ [0.5ex] 
\hline
$2D$ & 55.89 & 53.80 & 52.67 & 59.72 & 59.10 & 62.35 \\ [0.5ex] 
\hline
$3D$ & 55.47 & 52.60 & 55.21 & 56.33 & 61.42 & \textbf{63.43} \\ [0.5ex] 
\hline
\end{tabular}
\end{center}
\caption{Classification rates in \% of KTH dataset (8-frame videos) over 3 runs, as a function of objective time values for three layers with 2D and 3D multi-layer architectures.}
\label{table:obbjtimesimilar}
\end{table}

\subsection{Convolutional kernel}
The convolutional kernel size has a direct effect on the learned features, so in this section we present the classification rates obtained using different convolutional kernel sizes with both 2D and 3D architectures. Tables \ref{table:convkern8} and \ref{table:convkern20} show the results obtained on KTH using 8 frames per video and 20 frames per video, respectively. In Table \ref{table:convkern8}, the classification rates obtained with a 3D convolutional SNN are slightly higher than those obtained with a 2D convolutional SNN architecture. The best classification rate is $63.43\%$, obtained with a 3D model that has a kernel size of $5 \times 5 \times 2$; this is only slightly higher than the classification rate of $62.35\%$ obtained with a 2D SNN that has a kernel size of $5 \times 5$. However, Table \ref{table:convkern20} shows that a larger video length decreases the performance of 2D SNNs. This is due to the information saturation that results from pooling the features extracted by the 2D SNN, which results in a sample that is harder to classify for the SVM than a sample made up of less frames. Therefore, we can deduce that 2D SNNs are limited because of information saturation. On the other hand, 3D convolution does not have this problem. Table \ref{table:convkern20} shows that 3D convolutional SNNs perform significantly better than 2D convolutional SNNs with longer video sequences. However, this significant increase is only spotted in the first two layers, and then there is a decrease in classification rate for the third layer. This behaviour suggests that the length of the video affects the learning in subsequent layers for 3D multi-layer architectures. For the rest of this paper, we use video samples made up of $8$ frames for 2D SNNs, and $20$ frames for 3D SNNs. The chosen kernel sizes are $5 \times 5$ and $5 \times 5 \times 2$ for the 2D and 3D SNNs respectively.

\begin{table}
\begin{center}
\begin{tabular}{| c  c  c  c |}
\hline
\rowcolor{aliceblue}  &  $f_{w} \times f_{h} \times f_{td}$ & \#Filters & L1 ~ L2 ~~ L3\\
\hline
\multirow{2}{0.25ex}{}$3D$  & $3\times 3 \times 2$ & $16 \times 32 \times 64$ & 54.63 56.33 56.33 \\ 
                            & $3 \times 3 \times 3$  & $16 \times 32 \times 64$ & 52.16 57.10 57.56 \\ 
\hline 
\multirow{2}{0.25ex}{}$3D$   & $5 \times 5 \times 2$ & $16 \times 32 \times 64$ & 56.33 61.42 \textbf{63.43} \\
                             & $5 \times 5 \times 3$ & $16 \times 32 \times 64$ & 55.32	62.73 62.04 \\ 
\hline
\multirow{2}{0.25ex}{}$3D$   & $7 \times 7 \times 2$ & $16 \times 32 \times 64$ & 55.40 60.80 56.94 \\ 
                             & $7 \times 7 \times 3$  & $16 \times 32 \times 64$ & 54.01 56.64 39.81 \\ 
\hline
\multirow{2}{0.25ex}{}$3D$  & $9 \times 9 \times 2$ & $16 \times 32 \times 64$ & 56.94 59.14 49.54 \\  
                            & $9 \times 9 \times 3$ & $16 \times 32 \times 64$ & 57.41 60.65 39.51 \\ 
\hline
$2D$  & $3 \times 3$ & $16 \times 32 \times 64$ & 48.38	48.15 47.22 \\ 
\hline
$2D$  & $5 \times 5$  & $16 \times 32 \times 64$ & 59.72 59.10	\textbf{62.35} \\ 
\hline
$2D$  & $7 \times 7$ & $16 \times 32 \times 64$ & 57.99 58.80	60.76 \\ 
\hline
$2D$  & $9 \times 9$ & $16 \times 32 \times 64$ & 54.78	59.10 41.20 \\ 
\hline
\end{tabular}
\end{center}
\caption{Classification rates in \% on the KTH dataset (8-frame videos) over 3 runs, as a function of different convolutional kernel sizes for 2D and 3D SNNs.}
\label{table:convkern8}
\end{table}

\begin{table}
\begin{center}
\begin{tabular}{| c  c  c  c |}
\hline
\rowcolor{aliceblue}  &  $f_{w} \times f_{h} \times f_{td}$ & \#Filters & L1 ~ L2 ~~ L3\\
\hline
\multirow{2}{0.25ex}{}$3D$   & $3\times 3 \times 2$ & $16 \times 32 \times 64$ & 65.59 67.59 63.27 \\ 
                             & $3 \times 3 \times 3$  & $16 \times 32 \times 64$ & 61.46 65.51 65.16 \\ 
\hline 
\multirow{2}{0.25ex}{}$3D$   & $5 \times 5 \times 2$ & $16 \times 32 \times 64$ & 62.19	\textbf{68.21} 63.12 \\   
                             & $5 \times 5 \times 3$ & $16 \times 32 \times 64$ & 62.04 67.59 64.35 \\ 
\hline
\multirow{2}{0.25ex}{}$3D$   & $7 \times 7 \times 2$ & $16 \times 32 \times 64$ & 62.19 67.59 59.26 \\ 
                             & $7 \times 7 \times 3$  & $16 \times 32 \times 64$ & 59.49 62.50 49.31 \\ 
\hline
\multirow{2}{0.25ex}{}$3D$   & $9 \times 9 \times 2$ & $16 \times 32 \times 64$ & 60.19	62.96 38.66 \\  
                             & $9 \times 9 \times 3$ & $16 \times 32 \times 64$ & 62.96	65.74 49.31 \\ 
\hline
$2D$  & $3 \times 3$ & $16 \times 32 \times 64$ & 45.37 51.39 50.93 \\ 
\hline
$2D$  & $5 \times 5$  & $16 \times 32 \times 64$ & 54.63 \textbf{58.80} 50.93 \\ 
\hline
$2D$  & $7 \times 7$ & $16 \times 32 \times 64$ & 52.31 56.94 44.91 \\ 
\hline
$2D$  & $9 \times 9$ & $16 \times 32 \times 64$ & 55.09 54.63 28.70 \\ 
\hline
\end{tabular}
\end{center}
\caption{Classification rates in \% on the KTH dataset (20-frame videos) over 3 runs, as a function of different convolutional kernel sizes for 2D and 3D SNNs.}
\label{table:convkern20}
\end{table}

\subsection{Comparison of the learned features}
\begin{figure}
\centerline{\includegraphics[scale=0.4]{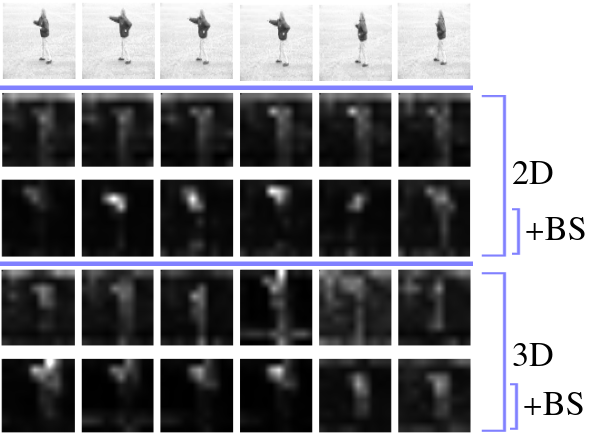}}
\caption{The feature maps in the first layer of the 2D and 3D SNN models from a KTH boxing video, with and without background subtraction. 1) The raw video frames, 2) feature maps with 2D SNN, 3) feature maps with 2D SNNs + background subtraction, 4) feature maps with 3D SNNs, and 5) feature maps with 3D SNNs + background subtraction.}
\label{fig:features}
\end{figure}

\begin{figure}
\centerline{\includegraphics[scale=0.4]{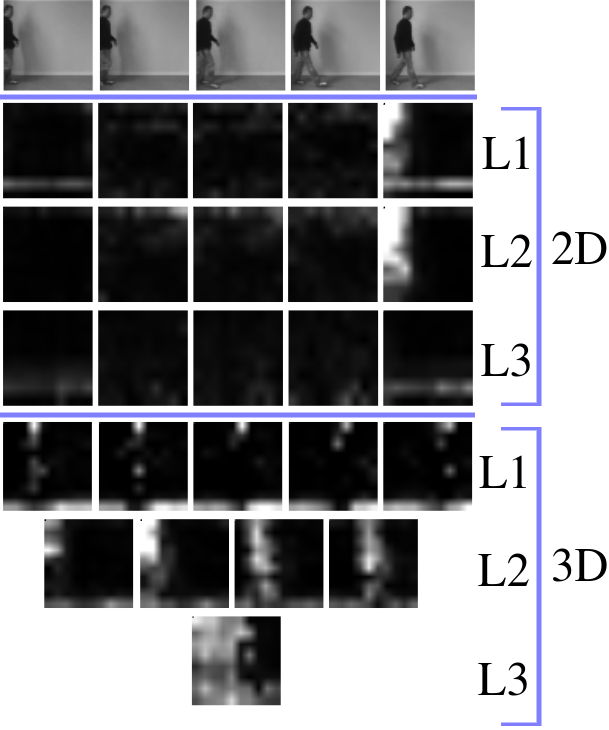}}
\caption{The feature maps of the 2D and 3D multi-layer SNN models from a KTH walking video.}
\label{fig:features2}
\end{figure}

The results of Table \ref{table:rawres} show that 3D convolutional networks yield better results with spatio-temporal information than 2D methods in all cases. However, the classification rate is decreasing again at the third layer, which would require further investigation. Moreover, it is interesting to study the behaviour of these networks with motion information as input. In Table \ref{table:spres}, we see the results of challenging these architectures with the motion information obtained by using background subtraction on the KTH and Weizmann datasets. The motion information has improved the classification rates with both the 2D and 3D architectures, so it is interesting to compare the feature maps provided by these models with and without background subtraction. In Figure \ref{fig:features}, comparing the feature maps extracted when the input video is pre-processed with background subtraction (+BS) shows that background subtraction has removed the fixed parts of the subjects body from the learned features, and highlights the moving parts of the subject (i.e. the hand in this boxing action). Therefore, the improvement in the classification rate is due to the network learning only the movement that is significant to identify an action; however, background subtraction may be less relevant when motion is subtle or when appearance is needed in classifying the action (e.g., when objects are involved). The feature maps obtained by the 2D SNN and the 3D SNN in Figure \ref{fig:features} seem similar. This is because this specific sample has been classified correctly by both architectures. Figure \ref{fig:features2} shows the feature maps obtained by 2D and 3D architectures for a walking sample, where this sample has been classified incorrectly by the 2D SNN, but has been classified correctly by the 3D SNN. We see that the 2D SNN failed to learn features that are significant enough to classify the sample from spatial information. The 3D SNN was able to learn and focus on motion information that enabled it to classify the walking action correctly. This proves that 3D SNNs trained with STDP can learn spatio-temporal features that are relevant. This also highlights the importance of convolution in the temporal dimension during spatio-temporal feature extraction with spiking models.

\begin{table}
\begin{center}
\begin{tabular}{|c | c | c | c | c | c | c |}  
\hline
\rowcolor{aliceblue}Dataset  & \multicolumn{3}{c|}{KTH} & \multicolumn{3}{c|}{Weizmann} \\
\hline
\rowcolor{aliceblue}Layers & Cnv1 & Cnv2 & Cnv3 & Cnv1 & Cnv2 & Cnv3 \\ [0.5ex] 
\hline
$2D$ & 57.78 & 58.61 & 58.80 & 48.12 & 55.38 & 55.98 \\ [0.5ex] 
\hline
$3D$ & 60.80 & \textbf{67.90}  & 64.20 & 52.96 & \textbf{60.55} & 57.92 \\ [0.5ex] 
\hline
\end{tabular}
\end{center}
\caption{Classification rates in \% with 2D and 3D multi-layer SNNs challenged with the KTH and Weizmann datasets over 3 runs.}
\label{table:rawres}
\end{table}

\begin{table}
\begin{center}
\begin{tabular}{|c | c | c | c | c | c | c |}  
\hline
\rowcolor{aliceblue}Dataset  & \multicolumn{3}{c|}{KTH} & \multicolumn{3}{c|}{Weizmann} \\
\hline
\rowcolor{aliceblue}Layers & Cnv1 & Cnv2 & Cnv3 & Cnv1 & Cnv2 & Cnv3 \\ [0.5ex] 
\hline
$2D$ & 61.57 & 61.11 & 61.11 & 60.43 & 61.28 & 62.39 \\ [0.5ex] 
\hline
$3D$ & 69.75 & \textbf{72.53}  & 66.05 & 61.28 & 61.54 & \textbf{64.62} \\ [0.5ex] 
\hline
\end{tabular}
\end{center}
\caption{Classification rates in \% with 2D and 3D multi-layer SNNs challenged with the motion information of the KTH and Weizmann datasets over 3 runs.}
\label{table:spres}
\end{table}

\section{Conclusion}
This work introduces 3D convolution to STDP-based spiking neural networks that learn features for action recognition. We also give an assessment of 2D and 3D convolutional neural network architectures trained with STDP and challenged with action recognition datasets. The results of this assessment yield several conclusions. The first one is that SNNs with unsupervised STDP can perform action recognition with no pre-processing. The second conclusion is that the performance of unsupervised STDP-based SNNs is still far behind that of state-of-the-art CNNs. However, it should be noted that our features are learned without supervision. Yet, further research is still needed to improve their performance. The third conclusion is that 3D convolutional SNN architectures trained with STDP can learn space-time features and clearly outperform 2D architectures. Finally, the last conclusion is that using a multi-layer architectures requires an exhaustive search to find the suitable hyper-parameters that permit learning in subsequent layers. This opens the door for other research questions, like what methods can be used to set hyper-parameters automatically \cite{PLIFICCV}, and how to improve the extraction of relevant space-time features with an end-to-end SNN. The 3D spiking neural network model tested in this work serves as a good starting point in improving human action recognition with unsupervised STDP-based convolutional SNNs.

\section*{Acknowledgments}
This work has been partially funded by IRCICA (USR 3380) under the bio-inspired project, and by the Luxant-ANVI industrial chair (I-Site and M\'etropole Europ\'eene de Lille).

\bibliographystyle{unsrt}
\bibliography{CitationLibrary}
\vspace{12pt}
\color{red}
\end{document}